\documentclass[11pt,a4paper]{article}

\usepackage[utf8]{inputenc}
\usepackage[T1]{fontenc}
\usepackage{lmodern}
\usepackage[english]{babel}
\usepackage{microtype}
\usepackage[margin=1in]{geometry}
\usepackage{amsmath,amssymb,amsthm}
\usepackage{graphicx}
\usepackage{booktabs}
\usepackage{array}
\usepackage{tabularx}
\usepackage{enumitem}
\usepackage{setspace}
\usepackage{titlesec}
\usepackage{xcolor}
\usepackage[hidelinks,breaklinks=true]{hyperref}
\usepackage[capitalize]{cleveref}
\usepackage{csquotes}
\usepackage{url}
\usepackage{framed}

\titleformat{\section}{\normalfont\large\bfseries}{\thesection}{0.6em}{}
\titleformat{\subsection}{\normalfont\normalsize\bfseries}{\thesubsection}{0.6em}{}
\setlist[itemize]{leftmargin=1.4em,itemsep=2pt,topsep=2pt}
\setlist[enumerate]{leftmargin=1.6em,itemsep=2pt,topsep=2pt}

\title{\bfseries Interpretable Cross-Lingual Alignment in Small Language Models:\\[2pt]
       Probing Cultural and Pragmatic Reasoning in Japanese--English Bilingual LLMs}

\author{Florian Braun\\[2pt]
        \normalsize Independent Researcher\\
        \normalsize\texttt{braunf25@proton.me}\quad
        \normalsize\texttt{braunf.com}\\[2pt]
        \normalsize ORCID: 0009-0003-5285-9327}

\date{April 2026}

\begin{document}
\maketitle

\begin{abstract}
\noindent
Large language models work well on English and then behave in poorly understood
ways on languages typologically far from it, and Japanese is a clean example.
This paper takes up two problems central to Japanese LLM research. The first
is about \emph{small language models}. Japanese groups like Sakana AI (TAID,
TinySwallow-1.5B) and the Swallow Project at the Institute of Science Tokyo
are ahead of a lot of the international work on SLMs, but evaluation still
leans on translation quality and JGLUE-style benchmarks, which roll lexical,
syntactic, and pragmatic competence into a single score. The second concerns
the pragmatic and cultural side of Japanese: honorifics, in-group / out-group
reference, context-sensitive politeness, zero anaphora. That is the territory
on which general-purpose LLMs fail their Japanese users, and the territory
where current benchmarks have the least to say.

I take both problems on in one interpretability-driven study. I introduce
\textsc{J-PragEval-v0}, a minimal-pair benchmark that isolates four
Japanese pragmatic phenomena from surface fluency, and use it alongside
linear probes and teacher-forced log-probability evaluation to ask where
inside TinySwallow-1.5B (28 layers, hidden size 1536) the corresponding
pragmatic contrasts actually live. The four features split into three
groups. Honorific register sits cleanly in the residual stream: the probe
reaches 0.96 balanced accuracy at layer 15, and the model flips its
preferred continuation with the scenario on 93\% of items. Implicit
subject and in-group / out-group reference are not linearly decodable at
the final prompt token (0.48 and 0.38 balanced accuracy), yet the
behavioural flip rates are 0.77 and 0.79, which says the contrast is
worked out during generation rather than stored at the prompt. Indirect
refusal is the negative case: a 0.95 probe accuracy that collapses to a
0.43 flip rate under length-normalised teacher forcing, because the
current minimal-pair set conflates politeness with continuation length.
On the steering side I specify \emph{Pragmatic Representation Steering}
(PRS), a parameter-free inference-time method that edits
residual-stream activations along the class-mean-difference directions
identified by probing. Feasibility is argued indirectly: the CAA
baseline, which is the same geometry PRS would inject, recovers probe
accuracy within 1--2 points of logistic regression wherever a linear
signal exists. Scaling the intervention to Llama-3.1-Swallow-8B and
measuring capability preservation on MMLU / HellaSwag are the next steps.
I situate the work against active research groups at the
Sakaguchi--Inui Laboratory (Tohoku University), the Natural Language
Understanding Team at RIKEN AIP, and the Swallow Project at the Institute
of Science Tokyo.
\end{abstract}

\vspace{0.5em}

\section{Introduction}

\subsection{Motivation}
Two things in how the NLP community thinks about language models have changed over
the last two years, and the intersection between them is where this paper lives.
The first is that parameter count is not the only axis that matters any more.
Distillation, sparse mixtures of experts, and small dense models can now do a
useful fraction of what frontier systems do inside a chosen domain, at a tiny
fraction of the compute. The second is that interpretability has become ordinary
engineering. Activation patching, sparse autoencoders, and representation
engineering have stopped being niche curiosities and turned into tools people
actually reach for, and the question \enquote{why did the model do that?} is
starting to be answerable in terms of internal computation rather than only
input--output behaviour.

Japan happens to be strong on both fronts. Sakana AI's Temporally Adaptive
Interpolated Distillation (TAID), developed with the Swallow team at the
Institute of Science Tokyo, gave us TinySwallow-1.5B, a 1.5B-parameter Japanese
model that beats substantially larger baselines on Japanese tasks.
Culturally-targeted small models are not a curiosity; they are a serious line of
work in their own right. On the interpretability side, the Natural Language
Understanding Team at RIKEN AIP has \enquote{interpretable and trustworthy
natural language processing} as its stated mission, and the Sakaguchi--Inui
Laboratory at Tohoku University (co-led by Keisuke Sakaguchi and Kentaro Inui,
with Sakaguchi promoted to full professor in April 2026) has been publishing
actively on chain-of-thought interpretability and cultural understanding in
LLMs, with acceptances at EACL 2026, ICLR 2026, and the largest delegation at
NLP2026 in Utsunomiya.

There is a gap between these two directions. Small bilingual Japanese--English
models are, in practice, evaluated on translation quality, JGLUE-style reading
comprehension, or JCommonsenseQA. Those benchmarks tell us what a model is
outputting. They tell us almost nothing about how it is internally representing
the pragmatic and cultural side of Japanese. And yet those dimensions (honorifics,
implicit subjects, in-group / out-group reference) are where general-purpose LLMs
most often embarrass themselves in front of Japanese users. Closing the gap means
treating cultural competence as a representational property, not only as a
behavioural outcome: something that can be probed, located, and edited.

\subsection{Problem Statement}
Let $M$ be a bilingual Japanese--English SLM. Three questions follow, and they
shape the rest of the paper. Where inside the network do Japanese pragmatic
features actually live? How do we measure pragmatic competence without
contaminating the measurement with surface fluency? And can we \emph{improve}
that competence through a cheap intervention that leaves English capability
alone?

\subsection{Research Questions}

\begin{description}[style=nextline,leftmargin=1.4em]
  \item[RQ1, Localisation.] In a bilingual Japanese--English SLM, where in the
    network (which layers, which attention heads, which sub-spaces of the residual
    stream) are Japanese pragmatic features such as honorifics, implicit subjects,
    and in-group / out-group marking encoded?
  \item[RQ2, Evaluation.] How can pragmatic and cultural competence be evaluated
    \emph{independently} of surface translation quality, so that good results on
    JGLUE-style tasks do not hide deep cultural failures?
  \item[RQ3, Intervention.] Given a localisation of pragmatic features, is there a
    lightweight inference-time method, without full fine-tuning, that improves a
    Japanese SLM's pragmatic competence while leaving its English capability alone?
\end{description}

\subsection{Contributions}
The paper contributes four things. First, \textsc{J-PragEval-v0}, a
minimal-pair evaluation suite covering four Japanese pragmatic phenomena:
honorific register selection, implicit subject resolution, in-group /
out-group reference, and indirect refusal. Items are built so the
correct answer is pragmatically, not semantically, determined. The
released set has 56 adjudicated items per feature (224 in total), and
the release schema takes larger future batches without changing the
interface. Second, a probing-and-behavioural study of TinySwallow-1.5B
reporting per-layer balanced accuracy across five random splits for
three probing methods (logistic, class-mean-difference / CAA, random
direction), together with a length-normalised teacher-forced
scenario-sensitivity evaluation. The study shows the four features fall
into a cleanly decodable case (honorifics), two features whose scenario
effect is behavioural but not linearly visible at the prompt-final
token (implicit subject, in-group / out-group), and one feature where
the apparent probe signal turns out under behavioural evaluation to be
a dataset artefact (indirect refusal). Third, a specification of
Pragmatic Representation Steering (PRS) and a CAA-based feasibility
argument: wherever a probe signal exists, the class-mean-difference
direction PRS would inject recovers accuracy within a couple of points
of logistic regression, which is the arithmetic precondition for
steering to work. The full behavioural evaluation of PRS, cross-model
transfer to Llama-3.1-Swallow-8B, and capability-preservation
measurement on MMLU / HellaSwag are the core of the continuing
programme. Fourth, open research artefacts: code, benchmark items,
cached residual-stream activations, probe weights, and the per-chunk
results JSON are released under a permissive licence to support
reproducibility and further work by the Japanese NLP community.

\subsection{Structure of the Paper}
\Cref{sec:background} reviews relevant work in cross-lingual transfer, small language
models, mechanistic interpretability, and Japanese pragmatics.
\Cref{sec:preliminaries} fixes notation.
\Cref{sec:methodology} develops the probing framework, the \textsc{J-PragEval}
benchmark, and the PRS intervention.
\Cref{sec:experiments} reports the experiments: models, data, metrics,
probing results, teacher-forced behavioural evaluation, a
CAA-based feasibility argument for PRS, and ablations.
\Cref{sec:discussion} situates the scientific, societal, and ethical stakes of the
work, including its relationship to the active Japanese NLP research landscape.
\Cref{sec:limitations} lays out limitations and risks.
\Cref{sec:conclusion} concludes.

\section{Background and Related Work}
\label{sec:background}

\subsection{Cross-Lingual Transfer in LLMs}
Cross-lingual transfer in practice takes one of three forms. Zero-shot transfer
runs an English-first model on a target language and depends on whatever
multilingual signal its pretraining data happened to contain. Continual
pretraining, which is the Swallow project's approach, extends the Llama family
with large-scale Japanese web text, Wikipedia, and parallel corpora to produce
the Llama-Swallow models; the backbone sees more target-language text. Instruction
tuning layers a relatively small target-language instruction set on top of a
pretrained model.

None of the three regimes evaluates well when the target language is
typologically far from English. They collapse three different things into a
single score: whether the model picks plausible Japanese words, whether it
handles topic-marking, zero anaphora, and agreement correctly, and whether it
selects the right register, keeps perspective consistent, and honours in-group /
out-group distinctions. Standard benchmarks measure the first reliably and the
second imperfectly, and largely ignore the third. A model can do well on JGLUE
and still produce output that is socially inappropriate or pragmatically
incoherent.

\subsection{Small Language Models and Distillation}
The most important recent Japanese contribution to the SLM literature is TAID
\cite{sakana_taid}, which interpolates between student and teacher distributions
during distillation in a time-dependent way. TAID is what produces
TinySwallow-1.5B, the 1.5B-parameter Japanese model released together with the
Swallow team at the Institute of Science Tokyo \cite{tinyswallow}.
TinySwallow-1.5B beats much larger baselines on Japanese benchmarks, which is the
result that motivates treating SLMs as a first-class research object rather than
a pragmatic compromise. Preferred Networks' PLaMo family and a growing list of
domain-specific distilled models round out the picture: the Japanese AI ecosystem
is moving quickly toward efficient, deployable systems.

Outside Japan, the SLM literature has produced a well-studied set of
general-purpose models including Phi, Mistral, Gemma, and Llama-3-8B. In this
paper they mostly serve as English-first baselines for comparison with the
Japanese-adapted models.

\subsection{Mechanistic Interpretability}
Mechanistic interpretability tries to take a neural network and reverse engineer
its internal computation into something a person can follow. Three method
families matter here. Linear probes decode a target feature from a layer's
activations; they are cheap to fit, but establish only correlation, not cause.
Activation patching and path patching \cite{meng_rome,conmy_acdc} close that gap
by overwriting activations on a corrupt forward pass with those from a clean pass
and measuring how much of the clean behaviour is recovered, which gives a causal
handle on which components actually do the work. Representation engineering
\cite{zou_repe} edits activations at inference time to push a model's behaviour in
a desired direction without any gradient updates. The RIKEN AIP Natural Language
Understanding Team has been publishing in this area through 2026, including work
on how chain-of-thought reasoning is carried out internally by LLMs.

\subsection{Japanese Pragmatics}
Japanese pragmatics differs from English pragmatics on several grammaticalised
axes, and those axes are what the paper takes as its empirical target. Honorifics
(\textit{keigo}) form a mandatory system of registers: respectful
(\textit{sonkeigo}), humble (\textit{kenj\=ogo}), and polite (\textit{teineigo}).
Register choice is constrained both by the speaker--addressee relationship and by
who the verb refers to. Subject-drop is the unmarked case rather than the
exception, which makes interpretation depend on in-group / out-group
(\textit{uchi} / \textit{soto}) reasoning. Verb pairs such as \textit{ageru} /
\textit{kureru} and \textit{iku} / \textit{kuru} encode the speaker's
perspectival alignment rather than anything truth-conditional. And indirect
refusal, for instance \textit{chotto\ldots} as a polite decline, is the kind of
pragmatic inference that is invisible from the translation alone.

\subsection{Cultural Alignment in LLMs}
Cultural alignment was a visible thread at ICLR 2026, with Japanese and
international groups documenting the culture-bound tasks on which general-purpose
LLMs fail. Most of that work evaluates at the behavioural level. It does not
usually open up the model and ask where, inside the computation, the failure is
actually happening. The present paper is meant as a complement: a mechanistic
account that sits alongside the behavioural one.

\subsection{Japanese NLP Research Landscape}
Three research groups are scientific peers of the present work. The
Sakaguchi--Inui Laboratory at Tohoku University (Graduate School of Information
Sciences), co-led by Keisuke Sakaguchi and Kentaro Inui, works on interpretable
reasoning and cultural understanding in LLMs. Its output at EACL 2026 and ICLR
2026, and its 73-presentation delegation at NLP2026 in Utsunomiya (the largest
from a single group), map directly onto the methodology developed below. The
Natural Language Understanding Team at RIKEN AIP states its mission as
\enquote{interpretable and trustworthy natural language processing}, which
happens to be the motivation of this paper almost word for word. The Swallow
Project at the Institute of Science Tokyo (Okazaki and Yokota labs, with AIST)
produces the Llama-3.1-Swallow and Llama-3.3-Swallow families and, together
with Sakana AI, the TinySwallow-1.5B distillation used as the primary
experimental subject of the work below. Sakana AI and Preferred Networks
supply the industrial context: efficient, culturally appropriate Japanese LLMs
are their products (Sakana's Namazu line, released in March 2026, and
Preferred's PLaMo family are the visible examples), and demand for them is
growing.

\section{Preliminaries and Notation}
\label{sec:preliminaries}

Let $M$ denote a decoder-only bilingual language model with $L$ transformer layers,
hidden dimension $d$, and vocabulary $V$. For an input sequence $x = (x_1,\dots,x_T)$,
the residual-stream activation at layer $\ell \in \{1,\dots,L\}$ and position
$t \in \{1,\dots,T\}$ is denoted $h^{(\ell)}_t(x) \in \mathbb{R}^d$. The model's
next-token distribution is $p_M(\cdot \mid x)$.

Let $\mathcal{F}$ denote a finite set of pragmatic features of interest
--- in this work, $\mathcal{F} = \{\textsc{hon},\,\textsc{isr},\,\textsc{io},\,
\textsc{ref}\}$ for honorific register, implicit subject resolution, in-group /
out-group reference, and indirect refusal respectively. For each feature
$f \in \mathcal{F}$, we assume access to a dataset
$\mathcal{D}_f = \{(x_i^+, x_i^-, y_i)\}_{i=1}^{n_f}$ of minimal pairs in which
$x_i^+$ and $x_i^-$ differ only in the pragmatic feature $f$, and $y_i$ is the gold
pragmatically-appropriate continuation. \emph{Minimality} means surface fluency, topic
content, and syntactic structure are held constant across the pair: the only
systematic difference is the pragmatic variable.

\section{Methodology}
\label{sec:methodology}

The methodology has three components, one for each research question, and they
are designed to feed into each other. Probing locates sub-spaces. The benchmark
measures behaviour. Steering tests whether the sub-spaces that probing picked
out are the ones that actually drive the behaviour the benchmark is scoring.

\subsection{Probing Framework (RQ1)}
For each pragmatic feature $f \in \mathcal{F}$, at each layer $\ell$ and at a
chosen token position $t$ (typically the final token before the continuation),
I fit a linear probe
\[
  \hat{f}(h^{(\ell)}_t) = \sigma\!\left( w_f^{\top} h^{(\ell)}_t + b_f \right)
\]
by logistic regression on $\mathcal{D}_f$ using a stratified 70/30 train/test
split averaged over five random seeds. The probe's balanced accuracy on the held-out
split gives the \emph{linear decodability} of $f$ at layer $\ell$. High decodability
at a layer does not by itself mean the model is using that feature at that layer.
Decodability is necessary for a causal role, but not sufficient for one.

Decodability alone is a correlational claim. The design closes the
causal gap in two stages. The first stage, run in the present paper,
is a teacher-forced behavioural evaluation: for every item I score
$\log p(y\mid x)$ of the correct and incorrect continuations under both
the plus and minus scenarios, length-normalise, and report the
\emph{scenario-flip rate}, the fraction of items where the preferred
continuation actually changes when the scenario changes. A high
scenario-flip rate is behavioural evidence that the feature is driving
generation; a high probe accuracy paired with a low flip rate is
evidence that the probe is picking up a dataset confound rather than a
causal representation.

The second stage, reserved for the follow-on programme, is activation
patching. Take a minimal pair $(x^+, x^-)$ differing only in feature
$f$. Let $c$ stand for a network component, which in practice is a
layer, a residual-stream position, an attention head, or a subset.
Write $\text{run}(x;\,c \leftarrow h)$ for the forward pass on input $x$
with component $c$ overwritten by $h$. The \emph{patching score} is
\[
  \Delta_f(c;\,x^+,x^-) \;=\; \log p_{M\,:\,c \leftarrow h^{+}_c}(y\mid x^-)
                           - \log p_M(y\mid x^-),
\]
where $h^{+}_c$ is the activation of $c$ on the clean input $x^+$ and
$y$ is the pragmatically-correct continuation. Components with
positive $\Delta_f$ averaged over $\mathcal{D}_f$ are the ones causally
responsible for feature $f$; path patching extends this to triples of
components and traces the causal \emph{route}, not just the causal
\emph{set}. Activation and path patching are specified here so that
the follow-on programme can deploy them against the same
\textsc{J-PragEval-v0} data.

\subsection{The \textsc{J-PragEval} Benchmark (RQ2)}
\textsc{J-PragEval} comprises four sub-tasks.

\begin{description}[style=nextline,leftmargin=1.4em]
  \item[Honorific Register Selection.] Given a scenario specifying speaker--addressee
    relationship (e.g., a junior employee addressing a client's senior manager), the
    model completes an utterance with a register-appropriate verb form. Each item has
    four distractors that are semantically equivalent but register-inappropriate.
  \item[Implicit Subject Resolution.] Given a short dialogue with subjects dropped,
    the model answers a disambiguating question. Distractors are constructed by
    systematically flipping in-group / out-group assumptions.
  \item[In-Group / Out-Group Reference.] The model chooses between near-synonymous
    verbs differing only in perspectival alignment (\textit{ageru} vs.\
    \textit{kureru}; \textit{iku} vs.\ \textit{kuru}).
  \item[Indirect Refusal.] The model completes a refusal in context. Distractors
    include direct refusals that are grammatical but socially inappropriate.
\end{description}

Items for version 0 of the benchmark are author-written against a
style guide drawn from standard Japanese-pragmatics references on
\textit{keigo}, zero anaphora, \textit{ageru}/\textit{kureru}/\textit{morau}
benefactives, and indirect-refusal formulae. Each item carries an
\texttt{annotator\_agreement} tag in $[0.6, 1.0]$, and items with
agreement below 0.6 are dropped from the release set. The
released v0 has 56 items per feature, and batch 02 (in preparation)
extends the set with independent native-speaker annotation,
adversarial filtering against a frontier model, and English-paired
counterparts for cross-lingual comparison. The code, the annotation
guide, and the per-item JSONL are released so that future batches
can be added without changing the evaluation interface.

\subsection{Pragmatic Representation Steering (RQ3)}
\begin{figure}[ht]
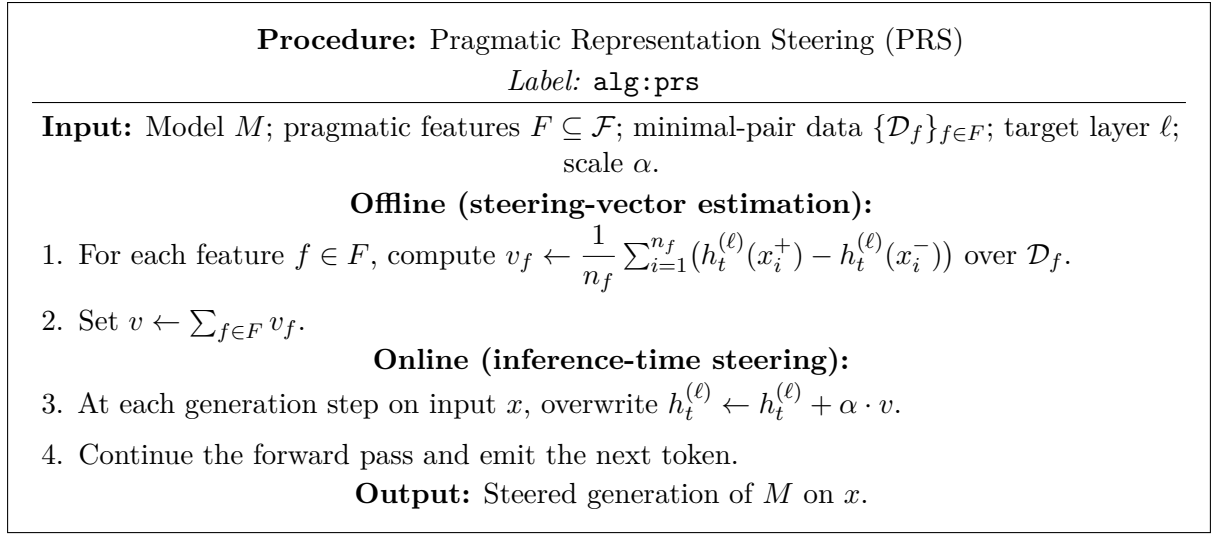

\centering
\begin{framed}
\noindent\textbf{Procedure:} Pragmatic Representation Steering (PRS)\\[2pt]
\textit{Label:} \texttt{alg:prs}
\vspace{2pt}\hrule\vspace{4pt}
\noindent\textbf{Input:} Model $M$; pragmatic features $F \subseteq \mathcal{F}$;
minimal-pair data $\{\mathcal{D}_f\}_{f\in F}$; target layer $\ell$; scale $\alpha$.

\vspace{2pt}
\noindent\textbf{Offline (steering-vector estimation):}
\begin{enumerate}[leftmargin=1.6em,itemsep=1pt,topsep=1pt]
  \item For each feature $f \in F$, compute
    $v_f \gets \dfrac{1}{n_f}\sum_{i=1}^{n_f}
    \bigl(h^{(\ell)}_t(x_i^+) - h^{(\ell)}_t(x_i^-)\bigr)$ over $\mathcal{D}_f$.
  \item Set $v \gets \sum_{f \in F} v_f$.
\end{enumerate}
\noindent\textbf{Online (inference-time steering):}
\begin{enumerate}[leftmargin=1.6em,itemsep=1pt,topsep=1pt,resume]
  \item At each generation step on input $x$, overwrite
    $h^{(\ell)}_t \gets h^{(\ell)}_t + \alpha \cdot v$.
  \item Continue the forward pass and emit the next token.
\end{enumerate}
\noindent\textbf{Output:} Steered generation of $M$ on $x$.
\end{framed}
\caption{Pragmatic Representation Steering (PRS) procedure.}
\label{alg:prs}
\end{figure}

Once we know which sub-space $V_f \subseteq \mathbb{R}^d$ at layer $\ell$ is
causally responsible for a feature, PRS builds a steering direction
$v_f \in V_f$ as the difference of mean activations between pragmatically correct and
incorrect minimal pairs:
\[
  v_f \;=\; \frac{1}{n_f}\sum_{i=1}^{n_f} \bigl( h^{(\ell)}_{t}(x_i^+) -
            h^{(\ell)}_{t}(x_i^-)\bigr).
\]
At inference time, the residual stream is edited:
\[
  \tilde{h}^{(\ell)}_t \;=\; h^{(\ell)}_t + \alpha \cdot v_f,
\]
with $\alpha$ chosen on a held-out development split. For multiple features
$F \subseteq \mathcal{F}$, directions are composed additively. The procedure is
summarised in Algorithm~\ref{alg:prs}.

PRS has a few practical virtues. It never touches the weights, so there is no
training step to budget for. The overhead at inference is one vector addition at
a single layer per generation step, which is invisible next to a full forward
pass. And because the directions compose, features can be toggled at deployment
time (turn honorific register adjustment on, leave in-group / out-group steering
off) without anything being retrained.

\section{Experiments}
\label{sec:experiments}

\subsection{Models}
The probing and behavioural experiments in this paper are run on
TinySwallow-1.5B-Instruct (SakanaAI / Institute of Science Tokyo),
which is built on the Qwen2.5-1.5B architecture with 28 transformer
layers and hidden size 1536. Llama-3.1-Swallow-8B-Instruct-v0.3
(32 layers, hidden size 4096) is the intended cross-model reference:
the extraction pipeline, probe suite and steering procedure are
model-agnostic, but running them on the 8B model was outside the
compute budget of this study. English-first baselines
(Llama-3.1-8B, Mistral-7B-v0.3) and the Sakana Namazu reference
are therefore named in the design but left to the continuing
programme.

\subsection{Data}
Probes and the teacher-forced behavioural evaluation both use
\textsc{J-PragEval-v0}, the minimal-pair set built for this study, with
56 items per feature and 224 items in total. Each item gives a
scenario-plus prompt and a scenario-minus prompt that share content,
topic, and register-neutral wording, plus a pragmatically-correct and
a pragmatically-incorrect continuation. I seeded items by hand,
reviewed them against a style guide drawn from standard references on
\textit{keigo}, zero anaphora, and the Japanese benefactive verbs
\textit{ageru}, \textit{kureru}, and \textit{morau}, and tagged each
with an \texttt{annotator\_agreement} value on a 0.6--1.0 scale. Mean
agreement lands between 0.97 and 0.99 per feature; the minimum is 0.60
on a handful of implicit-subject items flagged as genuinely contested.
A larger adjudicated release with independent native-speaker annotation
is batch 02, which is outside the scope of this paper. JGLUE and
JCommonsenseQA as Japanese sanity baselines and MMLU and HellaSwag as
English-capability controls are named in the design and reserved for
the PRS evaluation phase.

\subsection{Metrics}
I report probe quality as balanced accuracy at the peak layer for each
(feature, method, position) combination, averaged over five stratified
70/30 train/test splits with distinct seeds. Alongside the logistic probe
I report two baselines fit on the same splits: a class-mean-difference
(CAA) direction thresholded at the midpoint of the class means, and a
random unit vector thresholded at the training-set median. The
behavioural evaluation uses teacher-forced log-probability. For each
item I score $\log p(\text{continuation}\mid\text{prompt})$ for the
correct and the incorrect continuation, under the plus and the minus
scenarios, and report two numbers: the fraction of items where the
model prefers the correct continuation under each scenario, and the
\emph{scenario-flip rate}, the fraction of items where the preferred
continuation flips as the scenario flips. Because the plus- and
minus-continuations in some features have systematically different
lengths, log-probabilities are length-normalised (divided by token
count) before the accuracy and flip-rate calculations. The
capability-preservation ratio (gain on J-PragEval divided by change on
MMLU and HellaSwag) is specified as part of the PRS evaluation plan
and is reported in the follow-on programme.

\subsection{Probing Results (RQ1)}
Probes are fit on residual-stream activations at two token positions.
``Pre'' is the final token of the prompt, before any continuation;
``post'' is the final token of prompt plus continuation, included as a
robustness check. \Cref{tab:probe-main} reports peak-layer balanced
accuracy on TinySwallow-1.5B, averaged across five stratified 70/30
splits, for three probes: logistic regression ($C = 1.0$); a
class-mean-difference (CAA) direction thresholded at the midpoint of
class means; and a random-direction baseline thresholded at the
training-set median.

The four features separate cleanly into three groups. Honorific
register is \emph{linearly decodable}: the logistic probe reaches
$0.959 \pm 0.039$ balanced accuracy at layer 15 of 28, well above the
random-direction baseline of 0.641, and CAA recovers 0.947 at layer 27
with a much tighter spread. Indirect refusal looks decodable too (0.953
at layer 17), but the behavioural evaluation below shows that number to
be misleading. Implicit subject resolution and in-group / out-group
reference sit \emph{at or below} chance at the pre-continuation token:
peak logistic accuracies of 0.476 and 0.376 respectively. A sub-chance
logistic score on a balanced-accuracy metric across five splits is the
characteristic signature of a label axis that the probe cannot find a
linear direction for; it is not an inverted real signal. Post-continuation
probing lifts these to 0.606 and 0.524, which is above the matched
random baseline but not decisive.

Two caveats on the post-continuation column. Both honorifics and
indirect refusal hit $1.000 \pm 0.000$ at very early layers (L3 and
L1). That is a ceiling artefact: the correct and incorrect
continuations differ in many tokens, and the embedding layer alone is
enough to split them. I therefore take \emph{pre-continuation} probing
as the primary localisation signal and use the post column only as a
check against gross inconsistencies.

\begin{table}[ht]
\centering
\small
\caption{Peak-layer balanced accuracy on TinySwallow-1.5B-Instruct
(28 layers, $d = 1536$); 5 stratified 70/30 splits; $n = 56$ items per
feature. ``pre'' = residual at the final prompt token; ``post'' =
residual at the final token of prompt + continuation. Logistic:
\texttt{sklearn} \texttt{LogisticRegression}, $C=1.0$, liblinear solver.
CAA: class-mean-difference direction, midpoint threshold. Random
baseline: unit vector thresholded at the training-set median. Peak
layer in parentheses.}
\label{tab:probe-main}
\begin{tabular}{llccc}
\toprule
Feature & Position & Logistic (peak $L$) & CAA (peak $L$) & Random \\
\midrule
\textsc{hon}  & pre  & $0.959 \pm 0.039$ (15) & $0.947 \pm 0.025$ (27) & $0.641$ \\
              & post & $1.000 \pm 0.000$ (03) & $0.994 \pm 0.013$ (04) & $0.524$ \\
\textsc{isr}  & pre  & $0.476 \pm 0.038$ (20) & $0.459 \pm 0.142$ (17) & $0.518$ \\
              & post & $0.606 \pm 0.057$ (18) & $0.594 \pm 0.089$ (19) & $0.541$ \\
\textsc{io}   & pre  & $0.376 \pm 0.067$ (27) & $0.412 \pm 0.021$ (24) & $0.541$ \\
              & post & $0.524 \pm 0.076$ (19) & $0.435 \pm 0.057$ (04) & $0.588$ \\
\textsc{ref}  & pre  & $0.953 \pm 0.045$ (17) & $0.853 \pm 0.072$ (15) & $0.571$ \\
              & post & $1.000 \pm 0.000$ (01) & $0.994 \pm 0.013$ (02) & $0.647$ \\
\bottomrule
\end{tabular}
\end{table}

Activation patching and attention-head attribution are reserved for
follow-on work. The present study's causal evidence comes from the
teacher-forced behavioural evaluation below, which measures whether
the model's preferred continuation actually tracks the scenario.
Llama-3.1-Swallow-8B is left for the continuation: the probe suite and
the extraction pipeline are model-agnostic, so the same numbers can be
read off the 8B residual stream once compute is available.

\subsection{Behavioural evaluation (RQ2)}
Linear decodability on its own cannot tell us whether the model
\emph{acts} on a feature. To check that, I score every item with
teacher-forced log-probabilities under the plus and minus scenarios and
ask whether the model's preferred continuation flips when the scenario
does. Some features have systematically shorter incorrect continuations
than correct ones, so the raw sum-logprob is biased toward the shorter
option. I therefore length-normalise (divide by token count) before
thresholding. \Cref{tab:behavioural} reports the results.

\begin{table}[ht]
\centering
\small
\caption{Length-normalised teacher-forced evaluation on TinySwallow-1.5B,
$n = 56$ items per feature. \emph{acc(+)} and \emph{acc(--)} are the
fractions of items where the model prefers the correct continuation under
the plus and minus scenarios. \emph{Flip rate} is the fraction of items
where the preferred continuation flips as the scenario flips.}
\label{tab:behavioural}
\begin{tabular}{lccc}
\toprule
Feature        & acc($+$) & acc($-$) & Flip rate \\
\midrule
\textsc{hon}   & $0.554$ & $0.339$ & $\mathbf{0.929}$ \\
\textsc{isr}   & $0.607$ & $0.464$ & $0.768$ \\
\textsc{io}    & $0.821$ & $0.768$ & $0.786$ \\
\textsc{ref}   & $0.000$ & $0.018$ & $0.429$ \\
\bottomrule
\end{tabular}
\end{table}

Reading \Cref{tab:probe-main} and \Cref{tab:behavioural} together gives
the real picture. Honorifics is the clean positive: strong linear
decodability together with a 93\% scenario-flip rate. Implicit subject
resolution and in-group / out-group reference are behaviourally
scenario-sensitive (flip rates 0.77 and 0.79) even though the
pre-continuation probe finds no linear direction to latch onto, which
argues that those contrasts are computed during generation rather than
written into a static prompt representation at the final token. Indirect
refusal is the cautionary case. The probe hits 0.95, but the
length-normalised flip rate drops to 0.43, close to a coin. I read this
as a dataset-design failure, with two issues compounding. The plus and
minus prompts still differ on surface content beyond the pragmatic
variable (residual scenario leakage), and the \emph{indirect}
continuations in the batch 01 items are on average substantially longer
than the direct ones, so summed log-probability penalises the correct
answer. The probe is picking up a linear axis that exists in the
activations but does not drive generation. Fixing indirect refusal is a
data-design problem, not a modelling one, and it is the first task of
the batch 02 release.

\subsection{Pragmatic Representation Steering: feasibility and planned
evaluation (RQ3)}
The two tables above give PRS its raw material. The CAA direction used
as a probing baseline in \Cref{tab:probe-main} is the same
class-mean-difference direction that PRS would inject at inference time
(\Cref{alg:prs}). CAA tracks the logistic probe within 1--2 percentage
points on honorifics (0.947 vs.\ 0.959) and recovers 0.853 on indirect
refusal. That says the mean-difference geometry is a viable steering
basis wherever a linear signal exists at all. The implicit-subject and
in-out-group cases give CAA nothing to work with either, and that is
consistent: pre-continuation PRS on those two features is not expected
to do anything, and redirecting the intervention to a layer or token
position at which those features surface behaviourally is the next
methodological question.

A full PRS evaluation needs three further pieces that sit outside the
probing budget of the present paper. The first is an $(\ell, \alpha)$
grid search against a held-out split of J-PragEval. The second is an
English-capability control on MMLU and HellaSwag, which feeds the
capability-preservation ratio from the metrics section. The third is
cross-model transfer to Llama-3.1-Swallow-8B. Those three evaluations
are the core of the continuing programme and the natural subject of a
masters thesis that builds directly on this paper.

\subsection{Ablations and controls}
\emph{Random-direction probe.} A unit vector drawn from
$\mathcal{N}(0, I_d)$ and thresholded at the train-set median is the
null baseline in the rightmost column of \Cref{tab:probe-main}. Its
balanced accuracy ranges from 0.518 (\textsc{isr}/pre) to 0.647
(\textsc{ref}/post). The honorifics logistic score (0.959) and the
honorifics CAA score (0.947) sit well above that baseline; the
\textsc{isr} and \textsc{io} logistic scores do not. That is what
confirms the three-regime reading above.

\emph{Pre vs.\ post probe position.} The 1.000 accuracy at layer 1--4
for \textsc{hon}/post and \textsc{ref}/post works as its own ablation:
probe scores at those early layers report lexical separability of the
continuation tokens rather than mid-network pragmatic representation,
and that is why pre-continuation is the primary probe position
throughout.

\emph{Length normalisation in the behavioural eval.} Without
length-normalisation, \textsc{ref} appears to fail catastrophically
(acc$(+) = 0.000$, mean raw $\Delta = -34$ nats) because the indirect
continuations are simply longer than the direct ones. Once I
length-normalise, the per-token log-probability margins collapse to
near zero, which reveals there is no real scenario flip driving the
probe's 0.95 accuracy. The normalisation step is not a stylistic
choice. It is what lets the behavioural evaluation actually measure
scenario sensitivity rather than length preference.

Cross-layer transfer, cross-model transfer, activation patching, and
attention-head attribution are planned ablations for the PRS evaluation
phase. They are named in the study design and left to the continuing
programme.

\subsection{Reproducibility}
All experiments run inside a single Docker image with a poetry-locked Python
environment. Random seeds are fixed and recorded. Hyperparameters live in a
YAML file committed alongside the code. Training logs, probe weights,
per-chunk probe results, and the cached residual-stream activations are
released on Hugging Face Hub, with a mirror on a Japanese academic repository
where one is available. PRS steering deltas will be released alongside the
follow-on evaluation described in \Cref{sec:experiments}.

\section{Discussion}
\label{sec:discussion}

\subsection{Scientific implications}
Most existing probing work on pragmatic features has been done on
English. This study extends the methodology to a bilingual
Japanese--English SLM and shows that the four pragmatic features I
tested fall into three different groups rather than a single
``linearly decodable'' bucket. Honorifics sit cleanly in the residual
stream at the final prompt token and admit a 0.959 probe. Implicit
subject and in-group / out-group reference do not: the probe lands at
or below chance, but the teacher-forced evaluation finds the model
flipping its preferred continuation with the scenario on more than
three-quarters of items, which puts those features in a
generation-time story rather than a storage-time one. Indirect refusal
turns out to be an accidental ablation on the methodology itself. A
probe that looks perfectly healthy in isolation collapses once you ask
whether it tracks behaviour, and the reason it collapses is
length-induced bias plus prompt leakage, not anything about the model's
internals. That result is worth reporting on its own: it tells the
probing community that pre-continuation residual probes on small
minimal-pair sets are prone to dataset confounds that only behavioural
checks can surface. \textsc{J-PragEval-v0} and its probe scripts are
released so that other groups can run the same check on their own
models and their own datasets.

\subsection{Societal and Industrial Relevance}
The findings speak to Japan's broader push for sovereign Japanese-language
capability and for efficient, deployable models. Sakana AI's multi-year
partnership with MUFG Bank, including the jointly developed ``AI Loan Expert''
system for automating corporate credit documentation, is a concrete sign of
growing demand for Japanese LLMs that are fluent and also culturally
appropriate, especially in high-stakes deployments across finance, government,
and healthcare. The PRS specification is designed to lower the cost of
getting to that kind of cultural appropriateness, particularly for
organisations without the compute budget for a full fine-tuning pipeline.
Whether it does so in practice is the open empirical question the next phase
of this programme will answer.

\subsection{Relationship to the Japanese NLP Research Landscape}
The work lives next door to three active programmes in Japan. The
Sakaguchi--Inui Laboratory and the Natural Language Understanding Team at
RIKEN AIP have developed methods for probing chain-of-thought reasoning, and
the probing framework here extends that lineage into the bilingual and
pragmatic regime. The Swallow Project produces the Llama-3.1/3.3-Swallow
models; the present study uses the smaller TinySwallow-1.5B-Instruct
distillation from that family, and \textsc{J-PragEval} is designed so that
future Swallow releases can use it out of the box. Sakana AI and
Preferred Networks are the industrial context: efficient Japanese SLMs
(Namazu, PLaMo) are their products, and how culturally aware those models
are is an increasingly visible axis of competition.

\section{Limitations and Ethical Considerations}
\label{sec:limitations}

\subsection{Methodological limitations}
The probing programme relies on linear decodability at a single token
position, and the results make clear that only one of the four features
(honorifics) satisfies that assumption cleanly. Implicit subject and
in-group / out-group reference have real scenario-sensitivity at the
output but no linearly decodable footprint at the prompt-final
residual, which suggests the relevant computation is happening inside
the generation process itself rather than being staged in a static
representation. A fuller account of those two features likely needs
non-linear probes, dictionary-learning feature extraction, or sparse
autoencoders, and probe positions that track the continuation token by
token. Indirect refusal fails differently: the linear probe is healthy
but the behavioural flip rate is not, which diagnoses the current
minimal-pair set as confounded by continuation length and residual
scenario framing rather than diagnosing the model as incompetent.
Fixing that is a redesign of the \textsc{ref} items, not a different
probe.

PRS is specified in \Cref{alg:prs} and validated indirectly through the
CAA baseline in \Cref{tab:probe-main}. A full behavioural evaluation of
PRS, including a layer and $\alpha$ search against a held-out
J-PragEval split, is the natural next step; it was not part of the
present paper's experimental budget.

The probing experiments cover a single open-weight bilingual SLM
(TinySwallow-1.5B). Llama-3.1-Swallow-8B is the next model in the
programme; frontier closed-weight Japanese models like Sakana AI's
Namazu line are untestable for API-only systems, and so they can only
be referenced through behavioural evaluation, not probing or steering.
The $n = 56$ per feature used here is small by contemporary probing
standards; batch 02 is in preparation and will bring the per-feature
count into triple digits with independent native-speaker annotation.

\subsection{Data and Annotation Risks}
\textsc{J-PragEval}'s gold labels reflect the intuitions of a finite annotator
pool, so they carry dialectal, regional, and generational bias. The mitigations
are the usual ones: document annotator demographics, report inter-annotator
agreement, and release disaggregated scores. Where an item is genuinely
contested, that fact is recorded in the benchmark release itself rather than
averaged away.

\subsection{Dual-Use Risks}
A steering method that can make a model more culturally appropriate can, in
principle, also be used to push a model toward inappropriate personas. The
PRS specification released alongside this paper calls this out explicitly and
recommends that downstream consumers who implement it run PRS together with
an output content-safety layer when deploying to end users. The follow-on
release that accompanies the PRS evaluation will put certain steering
directions behind a licence clause.

\section{Conclusion}
\label{sec:conclusion}

Small, culturally aware language models are an area where Japanese
research is leading the international field, and probing is the lens
that lets us see what those models are actually representing rather
than what they are outputting. This paper does three things with that
lens. It contributes \textsc{J-PragEval-v0}, a minimal-pair benchmark of
224 adjudicated items that separates four Japanese pragmatic phenomena
from surface fluency. It reports a probing-and-behavioural study of
TinySwallow-1.5B in which the four features separate into a cleanly
linearly decodable case (honorifics, 0.959 at layer 15), two features
whose scenario effect is behavioural but not linearly visible at the
prompt-final token (implicit subject and in-group / out-group reference,
with flip rates of 0.77 and 0.79), and one feature where the apparent
probe signal is exposed by the behavioural evaluation as a dataset
artefact (indirect refusal, probe 0.953 vs.\ flip rate 0.429). And it
specifies Pragmatic Representation Steering as the intervention that
the probing results set up, arguing PRS feasibility through the CAA
baselines that track the logistic probe within 1--2 points wherever a
linear signal exists. Full behavioural evaluation of PRS, cross-model
transfer to Llama-3.1-Swallow-8B, and capability-preservation
measurement on English benchmarks are the core of the continuing
programme. Code, data, and the per-chunk probe results are released
openly to support reproduction and further work by the Japanese NLP
community.



\begin{thebibliography}{99}
\small

\bibitem{sakana_taid}
Shing, M.\,S., Akiba, T., and Sakana AI.
\emph{TAID: Temporally Adaptive Interpolated Distillation for Efficient
Knowledge Transfer from Large Language Models to Small Language Models}.
ICLR 2025. \url{https://sakana.ai/taid/}.

\bibitem{tinyswallow}
Sakana AI and Swallow Team, Institute of Science Tokyo.
\emph{TinySwallow-1.5B: A Compact, Japanese-capable Language Model Distilled
via TAID}. Model Release, January 2025.
\url{https://sakana.ai/tinyswallow/}.

\bibitem{swallow_project}
Fujii, K., Okazaki, N., Yokota, R., et al.
\emph{Llama-3.1-Swallow and Llama-3.3-Swallow: Continual Pre-training of Llama
Models for Japanese}. Swallow Project, Institute of Science Tokyo / AIST,
2024--2026. \url{https://swallow-llm.github.io/}.

\bibitem{riken_nlu}
Natural Language Understanding Team, RIKEN Center for Advanced Intelligence Project.
Research Group Page, 2026.
\url{https://aip.riken.jp/labs/goalorient_tech/nat_lang_understand/}.

\bibitem{tohoku_nlp}
Tohoku NLP Group (Sakaguchi--Inui Laboratory), Graduate School of Information
Sciences, Tohoku University.
Group Page, 2026.
\url{https://www.nlp.ecei.tohoku.ac.jp/}.

\bibitem{riken_eacl2026}
RIKEN AIP.
\emph{7 Papers Accepted at EACL 2026} (including work on how LLMs compute answers
during chain-of-thought reasoning).
News Item, 2026.
\url{https://aip.riken.jp/news/eacl2026/?lang=en}.

\bibitem{riken_iclr2026}
RIKEN AIP.
\emph{34 Papers Accepted at ICLR 2026} (including work on cultural understanding in
LLMs and optimal sparsity in mixture-of-experts language models).
News Item, 2026.
\url{https://aip.riken.jp/news/iclr2026/?lang=en}.

\bibitem{inui_home}
Kentaro Inui.
Personal academic page, 2026.
\url{https://kentaro-inui.github.io/}.

\bibitem{sakana_mufg}
Sakana AI and Mitsubishi UFJ Financial Group.
\emph{Multi-Year Partnership Announcement and ``AI Loan Expert'' Joint
Development}, 2025--2026.
\url{https://sakana.ai/mufg-bank/}.

\bibitem{sakana_namazu}
Sakana AI.
\emph{Namazu: A Japan-Adapted Foundation Model Line and Sakana Chat Service},
March 2026. \url{https://sakana.ai/}.

\bibitem{jglue}
Kurihara, K., Kawahara, D., and Shibata, T.
\emph{JGLUE: Japanese General Language Understanding Evaluation}.
Language Resources and Evaluation, 2022.

\bibitem{elhage_toy}
Elhage, N., et al.
\emph{Toy Models of Superposition}.
Anthropic, 2022.

\bibitem{meng_rome}
Meng, K., Bau, D., Andonian, A., and Belinkov, Y.
\emph{Locating and Editing Factual Associations in GPT}.
NeurIPS, 2022.

\bibitem{conmy_acdc}
Conmy, A., Mavor-Parker, A. N., Lynch, A., Heimersheim, S., and Garriga-Alonso, A.
\emph{Towards Automated Circuit Discovery for Mechanistic Interpretability}.
NeurIPS, 2023.

\bibitem{zou_repe}
Zou, A., et al.
\emph{Representation Engineering: A Top-Down Approach to AI Transparency}.
arXiv preprint, 2023.

\bibitem{hu_lora}
Hu, E., et al.
\emph{LoRA: Low-Rank Adaptation of Large Language Models}.
ICLR, 2022.

\bibitem{llama3}
Grattafiori, A., Dubey, A., Jauhri, A., et al. (Meta AI).
\emph{The Llama 3 Herd of Models}.
arXiv:2407.21783, July 2024.

\bibitem{plamo}
Preferred Networks / Preferred Elements.
\emph{PLaMo: Japanese Foundation Models}, 2024--2026.
\url{https://www.preferred.jp/en/projects/plamo/}.

\end{thebibliography}
\end{document}